\documentclass[11pt,a4paper]{article}
\usepackage[hyperref]{naaclhlt2019}
\usepackage{times}
\usepackage{booktabs}
\usepackage{graphicx}
\usepackage{amsmath}
\usepackage{pgfplots}
\usepackage{tikz}

\newcommand{\ignore}[1]{}

\aclfinalcopy

\newcommand*\samethanks[1][\value{footnote}]{\footnotemark[#1]}

\definecolor{g-blue}{HTML}{2E86C1}
\definecolor{g-red}{HTML}{B03A2E}
\definecolor{g-purple}{HTML}{AF7AC5}

\title{End-to-End Open-Domain Question Answering with BERTserini}

\author{Wei Yang,$^{1,2}$\thanks{\hspace{0.25cm}equal contribution} \hspace{0.1cm} Yuqing Xie,$^{1,2}$\samethanks \hspace{0.1cm}
Aileen Lin,$^2$
Xingyu Li,$^2$\\
{\bf Luchen Tan,}$^2$ {\bf Kun Xiong,}$^2$ {\bf Ming Li,}$^{1,2}$ \and {\bf Jimmy Lin}$^{1,2}$\vspace{0.1cm}\\
$^1$ David R. Cheriton School of Computer Science, University of Waterloo \\
$^2$ RSVP.ai}

\date{}

\begin{document}
\maketitle
\begin{abstract}
We demonstrate an end-to-end question answering system that integrates BERT with the open-source Anserini information retrieval toolkit.
In contrast to most question answering and reading comprehension models today, which operate over small amounts of input text, our system integrates best practices from IR with a BERT-based reader to identify answers from a large corpus of Wikipedia articles in an end-to-end fashion.
We report large improvements over previous results on a standard benchmark test collection, showing that fine-tuning pretrained BERT with SQuAD is sufficient to achieve high accuracy in identifying answer spans.
\end{abstract}

\section{Introduction}

BERT \cite{devlin2018bert}, the latest refinement of a series of neural models that make heavy use of pretraining \cite{N18-1202,radford2018improving}, has led to impressive gains in many natural language processing tasks, ranging from sentence classification to question answering to sequence labeling.
Most relevant to our task, \citet{nogueira2019passage} showed impressive gains in using BERT for query-based passage reranking.
In this demonstration, we integrate BERT with the open-source Anserini IR toolkit to create BERTserini, an end-to-end open-domain question answering (QA) system.

Unlike most QA or reading comprehension models, which are best described as rerankers or extractors since they assume as input relatively small amounts of text (an article, top $k$ sentences or passages, etc.), our system operates directly on a large corpus of Wikipedia articles.
We integrate best practices from the information retrieval community with BERT to produce an end-to-end system, and experiments on a standard benchmark test collection show large improvements over previous work.
Our results show that fine-tuning pretrained BERT with SQuAD~\cite{D16-1264} is sufficient to achieve high accuracy in identifying answer spans.
The simplicity of this design is one major feature of our architecture.
We have deployed BERTserini as a chatbot that users can interact with on diverse platforms, from laptops to mobile phones.

\section{Background and Related Work}

While the origins of question answering date back to the 1960s, the modern formulation can be traced to the Text Retrieval Conferences (TRECs) in the late 1990s~\cite{Voorhees_Tice_TREC8}.
With roots in information retrieval, it was generally envisioned that a QA system would comprise pipeline stages that select increasingly finer-grained segments of text~\cite{Tellex_etal_SIGIR2003}:\ document retrieval to identify relevant documents from a large corpus, followed by passage ranking to identify text segments that contain answers, and finally answer extraction to identify the answer spans.

As NLP researchers became increasingly interested in QA, they placed greater emphasis on the later stages of the pipeline to emphasize various aspects of linguistic analysis.
Information retrieval techniques receded into the background and became altogether ignored.
Most popular QA benchmark datasets today---for example, TrecQA~\cite{Yao13answerextraction}, WikiQA~\cite{yang2015wikiqa}, and MSMARCO~\cite{nguyen2016ms}---are best characterized as answer selection tasks.	
That is, the system is given the question as well as a candidate list of sentences to choose from.
Of course, those candidates have to come from {\it somewhere}, but their source lies outside the problem formulation.
Similarly, reading comprehension datasets such as SQuAD~\cite{D16-1264} eschew retrieval entirely, since there is only a single document from which to extract answers.

In contrast, what we refer to as ``end-to-end'' question answering begins with a large corpus of documents.
Since it is impractical to apply inference exhaustively to all documents in a corpus with current models (mostly based on neural networks), this formulation necessarily requires some type of term-based retrieval technique to restrict the input text under consideration---and hence an architecture quite like the pipelined systems from over a decade ago.
Recently, there has been a resurgence of interest in this task, the most notable of which is Dr.QA~\citep{P17-1171}.
Other recent papers have examined the role of retrieval in this end-to-end formulation~\citep{wang2017r,D18-1055,D18-1053}, some of which have, in essence, rediscovered ideas from the late 1990s and early 2000s.

For a wide range of applications, researchers have recently demonstrated the effectiveness of neural models that have been pretrained on a language modeling task~\cite{N18-1202,radford2018improving}; BERT~\cite{devlin2018bert} is the latest refinement of this idea.
Our work tackles end-to-end question answering by combining BERT with Anserini, an IR toolkit built on top of the popular open-source Lucene search engine.
Anserini~\cite{Yang_etal_SIGIR2017,Yang_etal_JDIQ2018} represents recent efforts by researchers to bring academic IR into better alignment with the practice of building real-world search applications, where Lucene has become the {\it de facto} platform used in industry.
Through an emphasis on rigorous software engineering and regression testing for replicability, Anserini codifies IR best practices today.
Recently, \citet{Lin_SIGIRForum2018} showed that a well-tuned Anserini implementation of a query expansion model proposed over a decade ago still beats two recent neural models for document ranking.
Thus, BERT and Anserini represent solid foundations on which to build an end-to-end question answering system.

\section{System Architecture}

\begin{figure}[t]
\centering
  \includegraphics[width=3in]{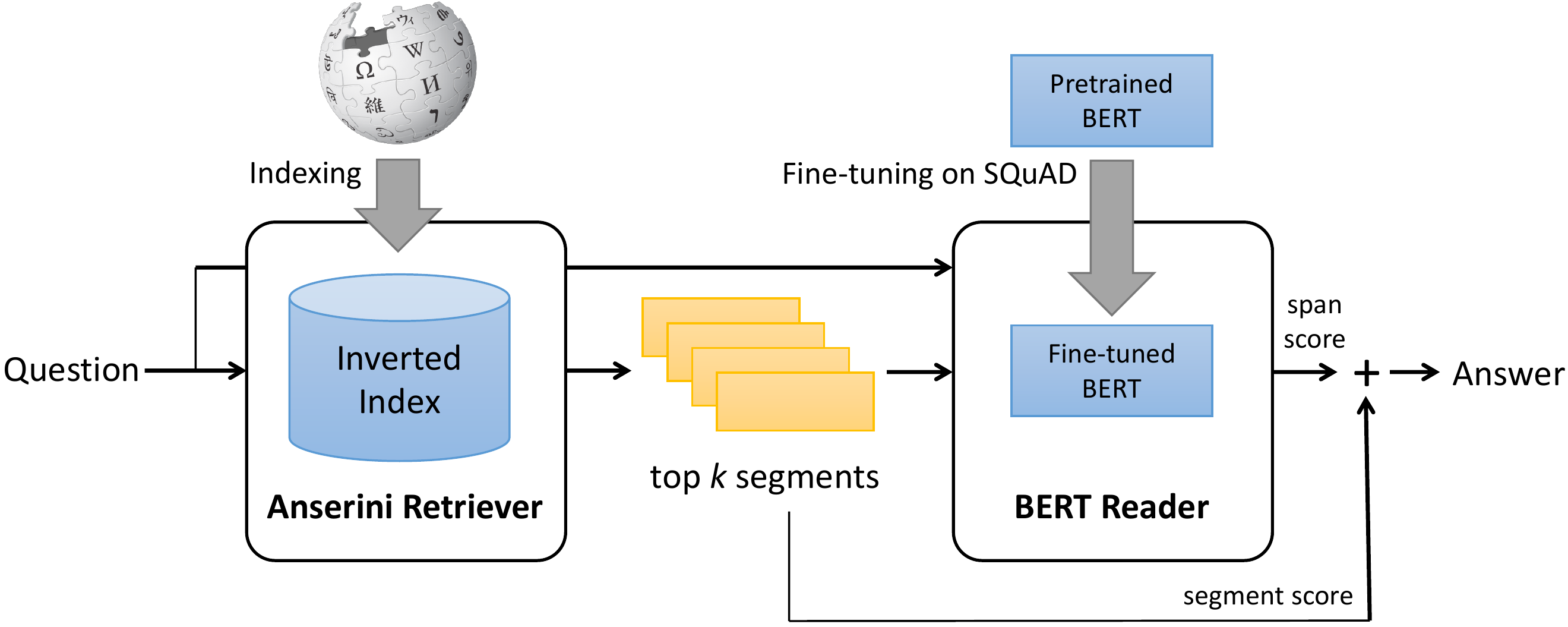}
\caption{Architecture of BERTserini.}
\label{fig:framework}
\end{figure}

The architecture of BERTserini is shown in Figure~\ref{fig:framework} and is comprised of two main modules, the Anserini retriever and the BERT reader.
The retriever is responsible for selecting segments of text that contain the answer, which is then passed to the reader to identify an answer span.
To facilitate comparisons to previous work, we use the same Wikipedia corpus described in~\citet{P17-1171} (from Dec.\ 2016) comprising 5.08M articles.
In what follows, we describe each module in turn.

\subsection{Anserini Retriever}

For simplicity, we adopted a single-stage retriever that directly identifies segments of text from Wikipedia to pass to the BERT reader---as opposed to a multi-stage retriever that first retrieves documents and then ranks passages within.
However, to increase flexibility, we experimented with different granularities of text at indexing time:

\smallskip \noindent \textbf{Article}:\ The 5.08M Wikipedia articles are directly indexed; that is, an article is the unit of retrieval.

\smallskip \noindent \textbf{Paragraph}:\ The corpus is pre-segmented into 29.5M paragraphs and indexed, where each paragraph is treated as a ``document'' (i.e., the unit of retrieval).

\smallskip \noindent \textbf{Sentence}:\ The corpus is pre-segmented into 79.5M sentences and indexed, where each sentence is treated as a ``document''.

\smallskip \noindent At inference time, we retrieve $k$ text segments (one of the above conditions) using the question as a ``bag of words'' query.
We use a post-v0.3.0 branch of Anserini,\footnote{\url{http://anserini.io/}} with BM25 as the ranking function (Anserini's default parameters).

\subsection{BERT Reader}

Text segments from the retriever are passed to the BERT reader.
We use the model in \citet{devlin2018bert}, but with one important difference:\ to allow comparison and aggregation of results from different segments, we remove the final softmax layer over different answer spans; cf.~\cite{P18-1078}.

Our BERT reader is based on Google's reference implementation\footnote{\url{https://github.com/google-research/bert}} (TensorFlow 1.12.0).
For training, we begin with the BERT-Base model (uncased, 12-layer, 768-hidden, 12-heads, 110M parameters) and then fine tune the model on the training set of SQuAD (v1.1).
All inputs to the reader are padded to 384 tokens; the learning rate is set to $3 \times 10^{-5}$ and all other defaults settings are used.

At inference time, for retrieved articles, we apply the BERT reader paragraph by paragraph.
For retrieved paragraphs, we apply inference over the entire paragraph.
For retrieved sentences, we apply inference over the entire sentence.
In all cases, the reader selects the best text span and provides a score.
We then combine the reader score with the retriever score via linear interpolation:
\begin{align*}
S = (1- \mu) \cdot S_{\textrm{Anserini}} + \mu \cdot S_{\textrm{BERT}}
\end{align*}
where $\mu \in [0,1]$ is a hyperparameter.
We tune $\mu$ on 1000 randomly-selected question-answer pairs from the SQuAD training set, considering all values in tenth increments.

\section{Experimental Results}

\begin{table}[t]
\centering\resizebox{\columnwidth}{!}{
\begin{tabular}{lccc}
\toprule
Model 				& EM   & F1   & R  \\ 
\midrule
Dr.QA \cite{P17-1171} 	        & 27.1 & -    & 77.8 \\
Dr.QA + Fine-tune 		& 28.4 & -    & -    \\
Dr.QA + Multitask 		& 29.8 & -    & -    \\
$\text{R}^3$ \cite{wang2017r} 	& 29.1 & 37.5 & -    \\
\citet{D18-1055} 		& 29.8 & -    & -    \\
Par. R. \cite{D18-1053}         & 28.5 & -    & 83.1 \\
Par. R. + Answer Agg. 	        & 28.9 & -    & -    \\
Par. R. + Full Agg.             & 30.2 & -    & -    \\
\textsc{Minimal}~\cite{P18-1160}         & 34.7 & 42.5 & 64.0 \\
\midrule
BERTserini (Article, $k=5$)     & 19.1 & 25.9 & 63.1 \\
BERTserini (Paragraph, $k=29$)  & 36.6 & 44.0 & 75.0 \\
BERTserini (Sentence, $k=78$)   & 34.0 & 41.0 & 67.5 \\
\midrule
BERTserini (Paragraph, $k=100$) & \textbf{38.6}	& \textbf{46.1} & \textbf{85.8} \\
\bottomrule
\end{tabular}}
\caption{Results on $\text{SQuAD}$ development questions.}
\label{table:main}
\end{table}

We adopt exactly the same evaluation methodology as \citet{P17-1171}, which was also used in subsequent work.
Test questions come from the development set of SQuAD; since our answers come from different texts, we only evaluate with respect to the SQuAD answer spans (i.e., the passage context is ignored).
Our evaluation metrics are also the same as \citet{P17-1171}:\ exact match (EM) score and F1 score (at the token level).
In addition, we compute recall (R), the fraction of questions for which the correct answer appears in {\it any} retrieved segment; this is what \citet{P17-1171} call the document retrieval results.
Note that this recall is {\it not} the same as the token-level recall component in the F1 score.

Our main results are shown in Table \ref{table:main}, where we report metrics with different Anserini retrieval conditions (article, paragraphs, and sentences).
We compare article retrieval at $k=5$, paragraph retrieval at $k=29$, and sentence retrieval at $k=78$.
The article setting matches the retrieval condition in \citet{P17-1171}.
The values of $k$ for the paragraph and sentence conditions are selected so that the reader considers approximately the same amount of text:\ each paragraph contains 2.7 sentences on average, and each article contains 5.8 paragraphs on average.
The table also copies results from previous work for comparison.

We see that article retrieval underperforms paragraph retrieval by a large margin:\ the reason, we believe, is that articles are long and contain many non-relevant sentences that serve as distractors to the BERT reader.
Sentences perform reasonably but not as well as paragraphs because they often lack the context for the reader to identify the answer span.
Paragraphs seem to represent a ``sweet spot'', yielding a large improvement in exact match score over previous results.

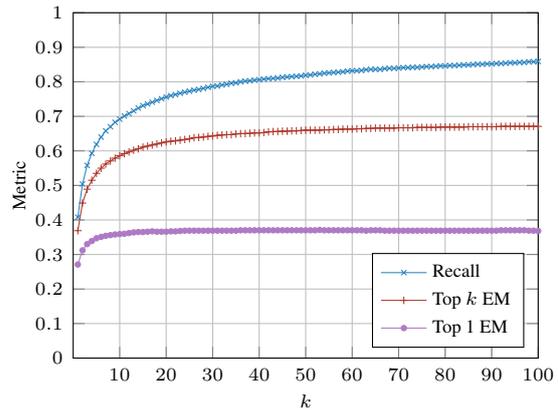
\begin{figure}[t]
\centering
\begin{tikzpicture}
\begin{axis}[
width=1.0\columnwidth,
height=0.8\columnwidth,
legend cell align=left,
legend style={at={(1, -0.1)},anchor=south east,font=\scriptsize},
mark options={mark size=3},
font=\scriptsize,
xmin=0, xmax=100,
ymin=0, ymax=1,
xtick={10, 20, 30, 40, 50, 60, 70, 80, 90, 100},
ytick={0,0.1, 0.2, 0.3, 0.4, 0.5, 0.6, 0.7, 0.8, 0.9, 1},
legend pos=south east,
xmajorgrids=true,
ymajorgrids=true,
xlabel style={yshift=1ex},
xlabel=$k$,
ylabel=Metric,
ylabel style={yshift=-3ex}
]    legend entries={Recall,
        Top k EM,
        Top 1 EM},
    ]
    \addplot[mark=x,g-blue, mark options={scale=0.7}] plot coordinates {
        (1, 0.408)(2, 0.504)(3, 0.558)(4, 0.593)(5, 0.619)(6, 0.64)(7, 0.658)(8,
0.67)(9, 0.683)(10, 0.692)(11, 0.701)(12, 0.708)(13, 0.716)(14, 0.724)(15,
0.731)(16, 0.736)(17, 0.741)(18, 0.746)(19, 0.751)(20, 0.756)(21, 0.76)(22,
0.763)(23, 0.767)(24, 0.77)(25, 0.773)(26, 0.776)(27, 0.778)(28, 0.781)(29,
0.784)(30, 0.787)(31, 0.788)(32, 0.791)(33, 0.793)(34, 0.796)(35, 0.797)(36,
0.8)(37, 0.802)(38, 0.803)(39, 0.805)(40, 0.806)(41, 0.808)(42, 0.809)(43,
0.81)(44, 0.811)(45, 0.813)(46, 0.814)(47, 0.815)(48, 0.816)(49, 0.817)(50,
0.819)(51, 0.82)(52, 0.822)(53, 0.823)(54, 0.825)(55, 0.826)(56, 0.827)(57,
0.828)(58, 0.829)(59, 0.831)(60, 0.832)(61, 0.832)(62, 0.833)(63, 0.835)(64,
0.836)(65, 0.836)(66, 0.836)(67, 0.838)(68, 0.839)(69, 0.839)(70, 0.84)(71,
0.841)(72, 0.841)(73, 0.842)(74, 0.842)(75, 0.843)(76, 0.844)(77, 0.844)(78,
0.845)(79, 0.846)(80, 0.846)(81, 0.847)(82, 0.847)(83, 0.848)(84, 0.849)(85,
0.849)(86, 0.85)(87, 0.851)(88, 0.851)(89, 0.852)(90, 0.852)(91, 0.853)(92,
0.853)(93, 0.854)(94, 0.855)(95, 0.855)(96, 0.856)(97, 0.856)(98, 0.858)(99,
0.858)(100, 0.859)
    };
    \addlegendentry{Recall}
    \addplot[mark=+,g-red, mark options={scale=0.7}] plot coordinates  {
    (1, 0.369)(2, 0.449)(3, 0.489)(4, 0.515)(5, 0.535)(6, 0.55)(7, 0.562)(8,
0.571)(9, 0.579)(10, 0.586)(11, 0.592)(12, 0.597)(13, 0.602)(14, 0.606)(15,
0.611)(16, 0.614)(17, 0.617)(18, 0.62)(19, 0.623)(20, 0.626)(21, 0.628)(22,
0.629)(23, 0.631)(24, 0.633)(25, 0.635)(26, 0.638)(27, 0.639)(28, 0.64)(29,
0.642)(30, 0.643)(31, 0.645)(32, 0.646)(33, 0.647)(34, 0.647)(35, 0.649)(36,
0.65)(37, 0.65)(38, 0.651)(39, 0.652)(40, 0.652)(41, 0.653)(42, 0.655)(43,
0.656)(44, 0.656)(45, 0.657)(46, 0.657)(47, 0.658)(48, 0.658)(49, 0.659)(50,
0.66)(51, 0.66)(52, 0.66)(53, 0.66)(54, 0.661)(55, 0.661)(56, 0.662)(57,
0.662)(58, 0.663)(59, 0.663)(60, 0.663)(61, 0.664)(62, 0.664)(63, 0.665)(64,
0.665)(65, 0.666)(66, 0.666)(67, 0.666)(68, 0.666)(69, 0.666)(70, 0.667)(71,
0.667)(72, 0.667)(73, 0.667)(74, 0.668)(75, 0.668)(76, 0.668)(77, 0.668)(78,
0.668)(79, 0.669)(80, 0.669)(81, 0.669)(82, 0.669)(83, 0.669)(84, 0.669)(85,
0.67)(86, 0.67)(87, 0.67)(88, 0.67)(89, 0.67)(90, 0.67)(91, 0.67)(92, 0.671)(93,
0.671)(94, 0.671)(95, 0.671)(96, 0.671)(97, 0.671)(98, 0.671)(99, 0.671)(100,
0.671)
    };
    \addlegendentry{Top $k$ EM}
    \addplot[mark=*,g-purple, mark options={scale=0.5}] plot coordinates {
   (1, 0.271)(2, 0.312)(3, 0.33)(4, 0.339)(5, 0.347)(6, 0.351)(7, 0.354)(8,
0.356)(9, 0.358)(10, 0.359)(11, 0.36)(12, 0.362)(13, 0.364)(14, 0.365)(15,
0.365)(16, 0.366)(17, 0.367)(18, 0.366)(19, 0.366)(20, 0.366)(21, 0.367)(22,
0.367)(23, 0.368)(24, 0.369)(25, 0.369)(26, 0.369)(27, 0.369)(28, 0.369)(29,
0.369)(30, 0.369)(31, 0.369)(32, 0.369)(33, 0.369)(34, 0.369)(35, 0.369)(36,
0.37)(37, 0.37)(38, 0.37)(39, 0.37)(40, 0.37)(41, 0.37)(42, 0.37)(43, 0.37)(44,
0.37)(45, 0.37)(46, 0.37)(47, 0.37)(48, 0.37)(49, 0.37)(50, 0.37)(51, 0.37)(52,
0.37)(53, 0.371)(54, 0.37)(55, 0.37)(56, 0.37)(57, 0.37)(58, 0.37)(59, 0.37)(60,
0.37)(61, 0.37)(62, 0.37)(63, 0.369)(64, 0.37)(65, 0.37)(66, 0.37)(67,
0.369)(68, 0.369)(69, 0.369)(70, 0.369)(71, 0.369)(72, 0.369)(73, 0.369)(74,
0.369)(75, 0.369)(76, 0.369)(77, 0.369)(78, 0.369)(79, 0.369)(80, 0.369)(81,
0.369)(82, 0.369)(83, 0.369)(84, 0.369)(85, 0.369)(86, 0.369)(87, 0.369)(88,
0.369)(89, 0.369)(90, 0.369)(91, 0.369)(92, 0.37)(93, 0.37)(94, 0.37)(95,
0.37)(96, 0.37)(97, 0.37)(98, 0.369)(99, 0.369)(100, 0.368)
    };
    \addlegendentry{Top 1 EM}
    \end{axis}
    
    \end{tikzpicture}
    \caption{Model effectiveness with different numbers of retrieved paragraphs.}
    \label{fig:pk}
\end{figure}

Our next experiment examined the effects of varying $k$, the number of text segments considered by the BERT reader.
Here, we focus only on the paragraph condition, with $\mu=0.5$ (the value learned via cross validation).
Figure~\ref{fig:pk} plots three metrics with respect to $k$:\ recall, top $k$ exact match, and top exact match.
Recall measures the fraction of questions for which the correct answer appears in {\it any} retrieved segment, exactly as in Table~\ref{table:main}.
Top $k$ exact match represents a lenient condition where the system receives credit for a correctly-identified span in {\it any} retrieved segment.
Finally, top exact match is evaluated with respect to the top-scoring span, comparable to the results reported in Table~\ref{table:main}.
Scores for the paragraph condition at $k=100$ are also reported in the table:\
we note that the exact match score is substantially higher than the previously-published best result that we are aware of.

We see that, as expected, scores increase with larger $k$ values.
However, the top exact match score doesn't appear to increase much after around $k = 10$.
The top $k$ exact match score continues growing a bit longer but also reaches saturation.
Recall appears to continue increasing all the way up to $k=100$, albeit more slowly as $k$ increases.
This means that the BERT reader is unable to take advantage of these additional answer passages that appear in the candidate pool.

These curves also provide a failure analysis:
The top recall curve (in blue) represents the upper bound with the current Anserini retriever.
At $k=100$, it is able to return at least one relevant paragraph around 86\% of the time, and thus we can conclude that passage retrieval does not appear to be the bottleneck in overall effectiveness in the current implementation.

The gap between the top blue recall curve and the top $k$ exact match curve (in red) quantifies the room for improvement with the BERT reader; these represent cases in which the reader did not identify the correct answer in {\it any} paragraph.
Finally, the gap between the red curve and the bottom top exact match curve (in purple) represents cases where BERT {\it did} identify the correct answer, but not as the top-scoring span.
This gap can be characterized as failures in scoring or score aggregation, and it seems to be the biggest area for improvement---suggesting that our current approach (weighted interpolation between the BERT and Anserini scores) is insufficient.
We are exploring reranking models that are capable of integrating more relevance signals.

One final caveat:\ this error analysis is based on the SQuAD ground truth.
Although our answers might not match the SQuAD answer spans, they may nevertheless be acceptable (for example, different answers to time-dependent questions).
In future work we plan on manually examining a sample of the errors to produce a more accurate classification of the failures.

\begin{figure}[t]
\vspace{0.15cm}
\centering\includegraphics[width=3in]{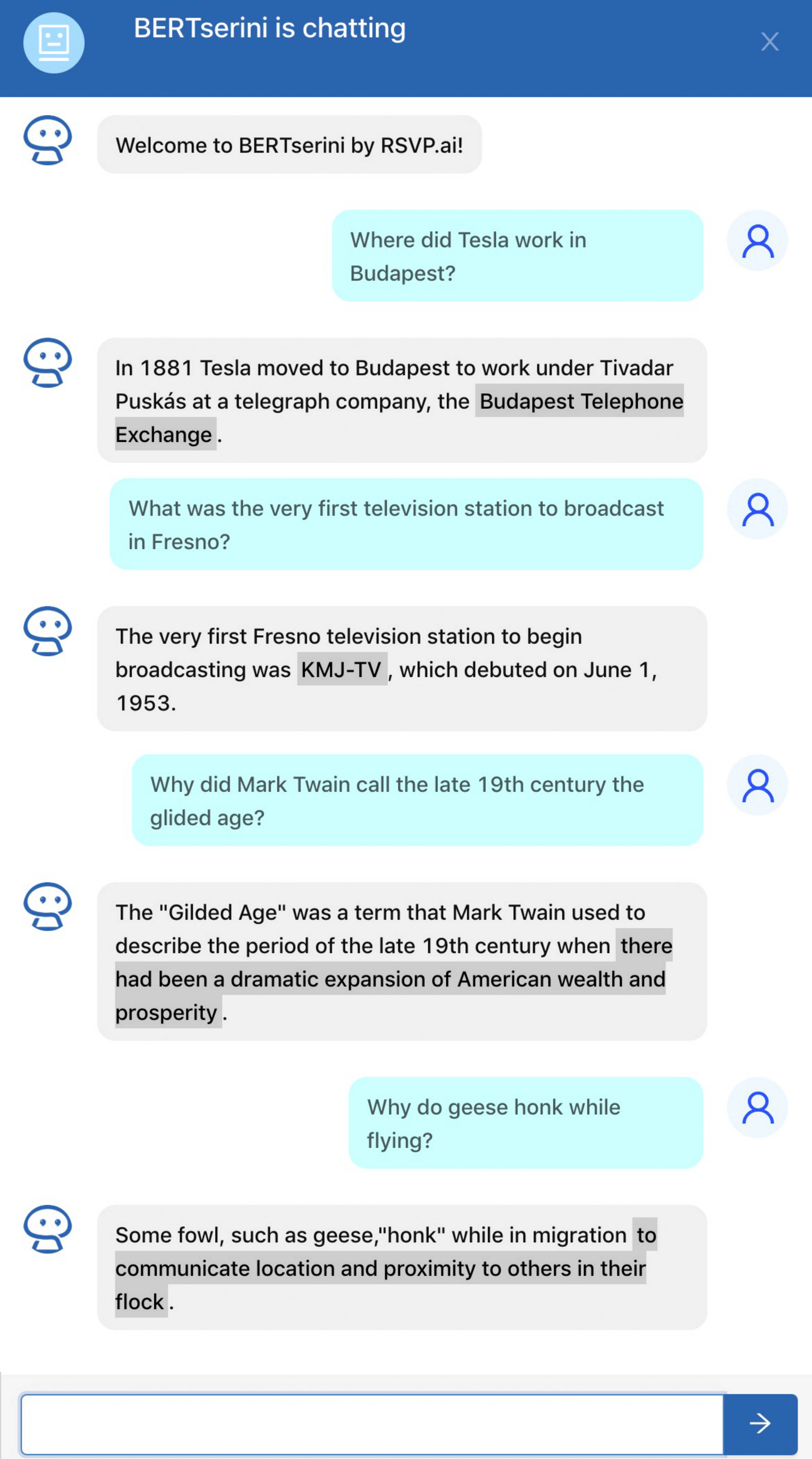}
\caption{A screenshot of BERTserini in RSVP.ai's chatbot interface. These samples from SQuAD illustrate the range of questions that the system can answer.}
\label{fig:demo}
\end{figure}

\section{Demonstration}

We have deployed BERTserini as a chatbot that users can interact with in two different ways:\ a Slackbot and RSVP.ai's intelligent platform that allows businesses to construct natural dialogue services easily and quickly.
However, both use the same backend services.
A screenshot of the RSVP.ai chat platform is shown in Figure \ref{fig:demo}.
The current interface uses the paragraph indexing condition, but we return only the sentence containing the answer identified by the BERT reader.
The answer span is highlighted in the response~\cite{Lin_etal_CHI2003}.
In the screenshot we can see the diversity of questions that BERTserini can handle---different types of named entities as well as queries whose answers are not noun phrases.

One important consideration in an operational system is the latency of the responses.
Informed by the analysis in Figure~\ref{fig:pk}, in our demonstration system we set $k=10$ under the paragraph condition.
While this does not give us the maximum possible accuracy, it represents a good cost/quality tradeoff.
To quantify processing time, we randomly selected 100 questions from SQuAD and recorded average latencies; measurements were taken on a machine with an Intel Xeon E5-2620 v4 CPU (2.10GHz) and a Tesla P40 GPU.
Anserini retrieval (on the CPU) averages 0.5s per question, while BERT processing time (on the GPU) averages 0.18s per question.

\section{Conclusion}

We introduce BERTserini, our end-to-end open-domain question answering system that integrates BERT and the Anserini IR toolkit.
With a simple two-stage pipeline architecture, we are able to achieve large improvements over previous systems.
Error analysis points to room for improvement in retrieval, answer extraction, and answer aggregation---all of which represent ongoing efforts.
In addition, we are also interested in expanding the multilingual capabilities of our system.

\bibliography{e2eqa-demo-naacl-final}
\bibliographystyle{acl_natbib}

\end{document}